\documentclass[10pt,journal,compsoc]{stylesheets/IEEEtran/IEEEtran}

\ifCLASSOPTIONcompsoc
  \usepackage[nocompress]{cite}
\else
  \usepackage{cite}
\fi
\ifCLASSINFOpdf
\else
\fi
\usepackage{url}

\usepackage[utf8]{inputenc}
\usepackage[textsize=scriptsize]{todonotes}
\usepackage{csquotes} 
\MakeOuterQuote{"} 
\usepackage{tikz}
\usetikzlibrary{positioning, calc, shapes, arrows}
\usepackage{amsmath}
\usepackage{amssymb}
\usepackage{mathtools}

\usepackage{algorithm}  
\usepackage{algorithmic}
\usepackage{booktabs}
\usepackage{multirow}
\usepackage{placeins} 
\usepackage{CJKutf8} 
\usepackage{bbding}
\usepackage[hidelinks]{hyperref}
\usepackage{stfloats}

\renewcommand{\sec}{\S}
\newcommand{\fig}{Fig.~}

\usepackage{rotating}

\begin{document}
%
\title{Emotion Embeddings --- \\
Learning Stable and Homogeneous Abstractions from Heterogeneous Affective Datasets
}
%
%
%
%

\author{Sven~Buechel~\IEEEmembership{}
        and~Udo~Hahn~\IEEEmembership{}
\IEEEcompsocitemizethanks{
\IEEEcompsocthanksitem 
S. Buechel and U. Hahn are with the Jena University Language and Information Engineering (JULIE) Lab, Friedrich Schiller University Jena
E-mail: firstname.lastname@uni-jena.de
}
\thanks{}
}

%
%

\markboth{Non-Archival Preprint}%
{Authors: Short-Title}
%



\IEEEtitleabstractindextext{%
\begin{abstract}
Human emotion is expressed in many communication modalities and media formats and so their computational study is equally diversified into natural language processing, audio signal analysis, computer vision, etc. Similarly, the large variety of representation formats used in previous research to describe emotions (polarity scales, basic emotion categories, dimensional approaches, appraisal theory, etc.) have led to an ever proliferating diversity of datasets, predictive models, and software tools for emotion analysis. Because of these two distinct types of heterogeneity, at the expressional and representational level, there is a dire need to unify previous work on increasingly diverging data and label types. This article presents such a unifying computational model. We propose a training procedure that learns a shared latent representation for emotions, so-called emotion embeddings, independent of different natural languages, communication modalities, media or representation label formats, and even disparate model architectures. Experiments on a wide range of heterogeneous affective datasets indicate that this approach yields the desired interoperability for the sake of reusability, interpretability and flexibility, without penalizing prediction quality. 
Code and data are archived under \href{https://doi.org/10.5281/zenodo.7405327}{DOI:10.5281/zenodo.7405327}.
\end{abstract}

}

\maketitle

\IEEEdisplaynontitleabstractindextext

%
\IEEEpeerreviewmaketitle


%
%
%
%


\section{Introduction} \label{sec:intro}

\begin{table*}[t!]
    \caption{Sample entries from various sources described along eight emotional variables
        \label{tab:examples}
    }
	\centering
	\setlength{\tabcolsep}{3pt}
    \newcommand{\lightcircle}{\raisebox{1mm}{$\circ$}}
\newcommand{\darkcircle}{\raisebox{1mm}{$\bullet$}}
\newcommand{\lighttriangle}{{\tiny\textsuperscript{$\triangle$}}} 
\newcommand{\lightsquare}{{\tiny\textsuperscript{$\square$}}}
\newcommand{\darksquare}{{\tiny\textsuperscript{$\blacksquare$}}}

	\begin{tabular}{l |ccc|ccccc cc}
	\toprule
	\textbf{Sample} &   \textbf{Valence} & \textbf{Arousal} & \textbf{Dominance} & \textbf{Joy} & \textbf{Anger} & \textbf{Sadness} & \textbf{Fear} & \textbf{Disgust} &  \textbf{Surprise}  & \textbf{Neutral}\\
    \midrule
		rollercoaster  	& 8.0\lightcircle & 8.1\lightcircle & 5.1\lightcircle & 3.4\lightsquare & 1.4\lightsquare	 & 1.1\lightsquare	& 2.8\lightsquare	& 1.1\lightsquare\\
		urine &	 3.3\lightcircle& 4.2\lightcircle & 5.2\lightcircle & 1.9\lightsquare & 1.4 \lightsquare & 1.2\lightsquare & 1.4\lightsquare &	2.6\lightsquare	 \\
		szcześliwy \textsuperscript{1}  & 2.8\darkcircle & 4.0\lightcircle & & &&& \\
		\midrule
		College tution continues climbing &&& &0\darksquare&54\darksquare & 40\darksquare & 3\darksquare & 31\darksquare & 3\darksquare\\ 
		A gentle, compassionate drama about grief...  & 1\lighttriangle &&&&&& \\ 
		\begin{CJK*}{UTF8}{bsmi}喇叭這一代還是差勁透了。\end{CJK*} \textsuperscript{2}  & 2.8\lightcircle & 6.1\lightcircle & & & & & &  \\ 
		\midrule
		\multirow{3}{*}{\raisebox{6pt}{\includegraphics[scale=.75]{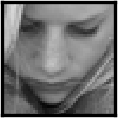}}    
		\raisebox{3pt}{\includegraphics[scale=.75]{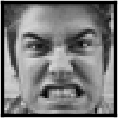}}
		\raisebox{0pt}{\includegraphics[scale=.75]{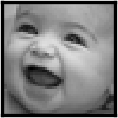}}} 
		&&&& 0\lighttriangle & 0\lighttriangle & 1\lighttriangle & 0\lighttriangle & 0\lighttriangle & 0\lighttriangle &  0\lighttriangle \\   
		&&& & 0\lighttriangle & 1\lighttriangle & 0\lighttriangle & 0\lighttriangle & 0\lighttriangle & 0\lighttriangle  & 0\lighttriangle \\
		& & & & 1\lighttriangle & 0\lighttriangle & 0\lighttriangle & 0\lighttriangle & 0\lighttriangle & 0\lighttriangle &  0\lighttriangle \\
		\bottomrule 
	   \multicolumn{8}{l}{\raisebox{-3pt}{Value Ranges: \hspace{1cm} \lightcircle $[1,9]$ \hspace{1cm} \darkcircle $[-3,3]$   \hspace{1cm} \lightsquare $[1, 5]$ \hspace{1cm} \darksquare $[0,100]$ \hspace{1cm}  \lighttriangle $\{1, 0\}$}} \\
	   \multicolumn{8}{l}{\raisebox{-3pt}{Translation:  \textsuperscript{1} Polish: \textit{happy};~ \textsuperscript{2} Chinese:\textit{ This product generation still has terrible speakers.}}}
	\end{tabular}

\end{table*}

In his famous quote, \textit{``You cannot} not \textit{communicate.''} \cite[p.52]{Watzlawick67}, Paul Watzlawick outlines the inevitability of human communication. In a similar vein, one might postulate that humans cannot avoid showing their own and absorbing others' emotions in their social activities. The ubiquity of emotions and their fundamental importance for human interaction justify the ample  efforts of a large and diverse research community ranging from social psychology and cognitive science up to the fields of artificial intelligence, machine learning (ML) and natural language processing (NLP). Yet, psychological and computational studies of emotion not only suffer from the enormous diversity of emotion phenotypes and modalities, but also from lots of often incompatible representation formats each leading to research niches whose results are hard to compare, if at all.
This woeful situation does not come as surprise when
one looks at the many varieties emotions can exhibit and how these can be described:

\textbf{Multi-Layered Composition of Emotions.}
Natural languages are among the primary carriers of emotion in social interaction. Their description is typically divided
into different axes (\textit{lexicon, syntax, semantics, discourse}) and analytic units (\textit{word, sentence, utterance in discourse}). This stratification is directly reflected in multi-layered linguistic analysis (or generation/production) steps where the emotional contributions of the most basic (lexical) level are passed on to higher levels of analysis comprising additional contextual information \cite{TangD16,JiaoW19}. This process is typically, but not necessarily, compositional in nature \cite{Yessenalina11,Socher13,ZhuX15}.

\textbf{Multi-Linguality of Emotions.} 
The term `natural language' abstracts away from thousands of \textit{specific instances of natural languages} (such as English, Mandarin or German), not to mention \textit{historic}, \textit{geographic}, or \textit{social variants} for each of them. All of these languages and their variants have (to some degree, at least) their own lexical, syntactic, semantic and pragmatic inventory for emotional expressions.

\textbf{Multi-Modality of Emotions.}  Verbal communication relies on two basic expression modes,  \textit{spoken} or \textit{written language}, i.e., we speak or listen, write or read. Each of these two modes provides a specialized repertoire of emotional signals supplementing the basic linguistic layers  (e.g., the lexicon) from above---spoken language, e.g., in terms of speech tone and rate, voice pitch or modulation, written language, e.g., in terms of the chosen register or variations in formality. Depending on the mode, additional modalities come into play \cite{WangY22,Ezzameli23}. For instance, in spoken conversations, facial expressions \cite{Navarretta16,ZhouF20,LeongS23}, head movements, body gestures, or other forms of non-verbal behavior \cite{Noroozi21,Mahfoudi22,LeongS23} often complement the basic spoken linguistic signals. In written communication, emoticons and emojis have become a common mood visualization system to complement basic written language signals \cite{KraljNovak15,Shoeb19,Illendula19,Hand22}.

\textbf{Multi-Mediality of Emotions.} Closely related to emotion modalities, yet more with emphasis on technical communication channels, different modes of emotion transmission can be distinguished \cite{ZhaoS19}. The \textit{auditory} channel is mostly served by spoken language for verbal communication \cite{Akcay20,SinghY22,Lope23}, but also sounds and music \cite{HanDonghong22,Panda23} are common carriers of emotions. The \textit{visual} channel is the primary one for the perception of written language, yet typically much more dominated by the perception of static \textit{visual objects}, such as paintings, photographs, images, cartoons, etc. \cite{ZhaoSicheng22}, which convey emotions in the same way as dynamic ones, such as videos, movies, or TV programs \cite{WangShangfei15,Abdu21}. In the dynamic case, audio and visual channels are combined in terms of audio-visual objects. 
As with the multi-layered analysis of emotions for natural languages, the combination of different media formats and the interactions among them (e.g., textual, audio and visual streams) is a more than challenging fusion problem for emotion analysis \cite{YangX21,ZhuLinan23}. Still, much less is known about the emotional role of haptic \cite{Eid16} and olfactory channels \cite{Calvi20}.

\textbf{Multiple Genres and Discourse Settings of Emotions.} The expression of emotions is also closely tied to the different communicative settings (\textit{genres}) in which they appear \cite{Chaffar11,Tafreshi18}. The (distribution of) emotions contained in classical mass media discourse (e.g., major headline news vs.\ commentaries), literary texts or essays differ markedly in tone and style from utterances spread via social media channels (e.g., in product or service reviews, chat groups or blogs) or the communication via e-mails \cite{Chhaya18}. The communicative habits and stylistic means (e.g., degrees of formality, profanity, irony, sarcasm, etc.) underlying these genres have a direct influence on the way emotional statements are to be interpreted genre-wise (this observation not only holds for textual data but also, e.g., for audio-visual (\cite{Muszynski21} for movies) or auditory ones (\cite{Eerola11,Griffiths21} 
for music)). 
Beyond genre, the \textit{discourse setting} is crucial for emotion analysis---written texts, with the exception of literary ones, typically assume a monologic scenario and by and large stable emotions (one speaker, the writer), whereas conversations are characterized by dialogues between two speakers or, as in multi-party discourse (discussions, debates, meetings), more than two speakers, where changes in emotions (so-called emotion shifts) are likely to occur as a result of the verbal interactions \cite{Poria19,GaoQ22}. 

In a similar vein, images differ in how they convey emotions---presumably, emotion is elicited in viewers via different mechanisms---depending on whether such images depict, say, oil paintings of landscape views or photographs of human faces or belong to different styles \cite{MohammadSM18}. 
Likewise, in the realm of audio objects, a recording of agitated speech and a piece of music (classical symphonies, jazz improvisations, or rap songs \cite{Imbir17}) differ in how and which emotions are communicated.

\textbf{Multiple Representation Formats of Emotions.} While the above-made distinctions mostly refer to the \textit{object level} of how emotions can be expressed via encoding procedures and interpreted via decoding procedures, the distinction of different representation formats addresses the \textit{meta-level} of emotion description. The study of emotions was initiated by scholars of (social) psychology and led to basically two streams of work \cite{Scherer00}. In the \textit{categorical approach}, basic emotion categories were stipulated (such as fear, anger, joy, sadness, etc.) and the emotional load of single emotion items was measured independently for each category, whereas in the \textit{dimensional approach} an n-dimensional emotion space is assumed and single emotional items are embedded as points into this space in a way that makes the geometrical interdependencies between the postulated emotion axes explicit.
Outside the psychological camp (mostly in the computer science community), a simple ternary division into three \textit{polarity stages} has found wide-spread use for sentiment analysis, namely positive, negative, and neutral sentiments.

These fundamental representational variants are currently reflected by an ever-increasing diversity of benchmarks and data resources (emotion lexicons, emotion-tagged corpora, etc.) that are often incompatible due to mutually exclusive representation formats for emotions (see Table \ref{tab:examples} for examples). Once the developers of an emotion analysis system have made their selection from these data resources (or created yet another new one) and representation formats, they often cannot be compared with alternative ones. 
While the total volume of available emotion data resources is truly impressive, they are, in effect, siloed based on specific choices of their modality, mediality, natural language and genre (levels of variability we subsequently summarize by the notion of \enquote{\textit{domain}}), as well as their choice of representation format.
Overall, this heterogeneity hampers progress for the field of affective computing.

To resolve this problem, in this article, we propose  \enquote{\textit{emotion embeddings}}, distributional representations of human affects that are both independent of the selection of the domain and its representation format, as well as stable across repeated executions of the generation process. In other words, samples that are linked to \textit{similar emotions}, irrespective of whether they are encoded in visual, auditory, or textual data, whether they are annotated with emotion dimensions or categories, will always receive \textit{similar embeddings}. As such, these embeddings form a common representational foundation for affective computing, tearing down the walls of the aforementioned data silos so that they form a single common ground for future research.

In more detail, this article describes how to learn these embeddings and how to further capitalize on them for typical inference problems and various application scenarios. On a technical level, our proposal consists of a sequence of steps where we first learn a model that can translate between multiple emotion label formats through a common intermediate representation---an "interlingua for emotion" (the emotion embeddings). This newly proposed \textit{multi-way mapping model} goes beyond previous work in its ability to translate between an arbitrary number of formats, whereas existing mapping models are fitted to one specific input and output format, respectively. Next, we propose a technique to use this model as a teacher to supervise the training process of further student models for various heterogeneous domains (images as well as texts and words in different languages). In doing so, the student models learn to embed their specific samples in the common emotion space	and, by extension, to infer ratings for a multitude of representation formats. To this end, we propose a new data augmentation strategy that exploits the diverse capabilities of our framework model to generate new training data on the fly. Our experiments, conducted on a total of 16 datasets, featuring facial images as well five different natural languages, demonstrate these benefits and confirm that they bear no negative impact on prediction quality.

The remainder of this article is structured as follows. \S\ref{sec:related} discusses related work and the novelty of our approach which is then detailed in \S\ref{sec:methods}. \S\ref{sec:setup} describes the setup of our experiments followed by their results in \S\ref{sec:results}. \S\ref{sec:analysis} gives further insights into our approach including ablation studies, visualizations of the learned emotion space, as well as a demonstration example for a novel application enabled by our proposal, viz. emotion-based retrieval. Finally, \S\ref{sec:discussion} discusses open issues and concludes this study.

\section{Related Work}
\label{sec:related}




\subsection{Representing Emotion}

At the heart of computational emotion representation lies a set of \textit{emotion variables}   used to capture different facets of affective meaning. Researchers may choose from a multitude of approaches developed in the long and controversial history of the psychology of emotion \cite{Scherer00,WangZhaoxia20}.  
Two fundamental streams of research can be distinguished here---categorical vs.\ dimensional approaches.
A very popular choice in the categorical school of thought are so-called \textit{basic emotions},
such as the six categories identified by Ekman \cite{Ekman92}: \textit{Joy}, \textit{Anger}, \textit{Sadness}, \textit{Fear}, \textit{Disgust}, and \textit{Surprise}. 
A subset of these excluding \textit{Surprise} is often used for emotional word datasets in psychology (\enquote{affective norms}). 
Other influential categorical systems include Plutchik's \textit{Wheel of Emotion}  \cite{Plutchik1980} and the componential approach underlying appraisal theories \cite{MoorsScherer13}.

\textit{Affective dimensions} constitute the most influential counterdraft to categorical conceptualizations of emotions in psychology \cite{Russell77,Bradley94,Mehrabian96}. 
The most important ones are \textit{Valence} (negative vs.\ positive, thus corresponding to the notion of \textit{polarity} \cite{Turney03}) and \textit{Arousal} (calm vs.\ excited). These two dimensions are  sometimes extended by \textit{Dominance} (feeling powerless vs.\ empowered).    

Inspired by these psychological theories of emotion the rapidly emerging field of affective computing started building on these foundations, incorporating both categorical 
as well as dimensional approaches.
Categorical models found use for textual (\cite{Mohammad12,Tafreshi18,Wang20tit,Hofmann20}), auditory (\cite{XieYue19,Burkhardt22} for speech, \cite{Eerola+V11,YangYH12} for music), and visual data (\cite{ChenYL17,Akhtar19} for videos, \cite{LiLiang19,Kosti20} for images, \cite{Illendula19} for emojis), as well as mixed media scenarios (see \cite{YangX21} for a combination of image and text data, \cite{Pandeya21} for music video data). 
Similarly, dimensional models were applied to textual (\cite{Buechel16ecai,WuC19,ChangYC19,Vishnubhotla22,Shah22aacl,Mendes23}), auditory (\cite{Dai15} for vocal social media, \cite{ZhangZ19,Atmaja20} for speech, \cite{Eerola+V11,Griffiths21} for music), visual data (\cite{WangX19,ZhouF20} for facial images, \cite{ZhaoS17,Kosti20} for images, \cite{Soleymani16} for videos incorporating neurophysiological signals and facial impressions of viewers, \cite{Guendil17} for music videos incorporating neurophysiological data), and a combination of audio-visual data \cite{Tzirakis17,NguyenD22}.

Yet, frequently, studies do not follow any of these established approaches but rather invent, more or less in an \textit{ad hoc} fashion, a customized set of variables, often reflecting the availability of user-labeled data in social media, the specifics of an application scenario, or the chosen domain which requires attention to particular emotional nuances (\cite{Abdul17acl,MohammadSM18,LiLiang19,XieYue19,Kosti20,Pandeya21,YangX21,Araque22}).

Besides the set of variables considered for representing emotion, another important decision relates to the values these variables may take, which in turn links to specific learning problems. In particular, allowing variables to take values from some real-valued interval translates to an emotion \textit{regression} problem, whereas allowing them to take only binary values translates into a \textit{classification} problem. In case of real-valued labels, datasets also differ in their specific value ranges, e.g., $[1,9]$, $[1,5]$, or $[0,1]$ (cf.\ Table \ref{tab:examples} for an illustration).\footnote{Weidman et al.\ \cite{Weidman17} also point at unsolved reliability issues arising from the use of different impromptu scales for the same emotion category.} Similarly, categorical datasets differ in whether only a single or multiple variables may be instantiated per item (single-label vs.\ multi-label classification). 

\begin{figure}[t]
    \centering
    \includegraphics[width=.5\textwidth]{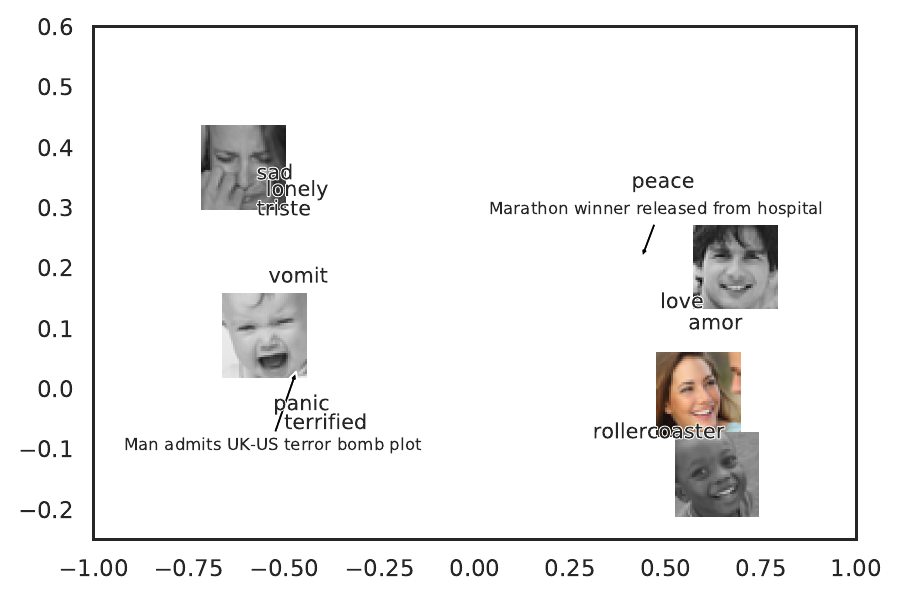}
    \caption{Emotional loading of heterogeneous samples in common representation space in first two principal components. 
    }
    \label{fig:emotion-space}
\end{figure}

This proliferating diversity of emotion label formats is the reason for the lack of comparability outlined in \sec \ref{sec:intro}. Our work aims to unify these heterogeneous labels by learning to translate them into a shared distributional representation (see \fig \ref{fig:emotion-space}).

\subsection{Analyzing Emotion}

There are several subtasks in emotion analysis that require distinct model types \cite{Sailunaz18}. In this study, we will focus on (1) single word items, (2) texts of varying length, and (3) static images. Other important input types, not further discussed here, are video (i.e., dynamic image sequences; \cite{WangShangfei15,Abdu21}), audio (speech (\cite{Swain18,Akcay20,SinghY22,Lope23,Latif23}) and non-speech sounds, such as music (\cite{Eerola+V11,HanDonghong22,Panda23})), and different kinds of bio-signals as physiological correlates of emotions, e.g.,  skin conductivity, muscle activities, EEG signals, or heart rate (\cite{Shukla21,GarciaMartinez21,Harper22}).

\textit{Word}-level prediction (\enquote{emotion lexicon induction}) is concerned with the emotion associated with an individual word \textit{out of context}. Early work primarily exploited surface patterns of word usage \cite{Hatzivassiloglou97acl,Turney03}, whereas more recent studies rely on more sophisticated statistical signals encoded in word embeddings \cite{Amir15,Rothe16,Li17}. Combinations of high-quality embeddings with feed-forward nets have shown high potential, rivaling human annotation capabilities \cite{Buechel18naacl}.

In contrast, modeling the emotion status \textit{within the context} of sentences or short texts (here jointly referred to as \textit{\enquote{text}}) \cite{Nandwani21,Deng23} was traditionally based largely on lexical resources \cite{Taboada11cl}. Later, those were combined with conventional feature-engineered machine learning techniques \cite{Mohammed13semeval}, before being widely replaced by neural end-to-end approaches 
\cite{Socher13,Abdul17acl}.
Current state-of-the-art results are achieved by transfer learning with transformer models \cite{Zhong19,Delbrouck20,Siriwardhana20,Acheampong21}.

Regarding the emotion capacity of \textit{image} data, there is a deep divide in the research traditions between facial emotion recognition (concerned with detecting the emotion depicted in an image of a human face \cite{Goodfellow15,Mollahosseini19,ZhouF20}) and visual emotion analysis (asking for the emotion induced by viewing a certain image \cite{ZhangWei20,XuL22,Saadon23}). 

Further focusing this overview on facial emotion recognition in computer vision \cite{LiS22,Jampour22,LeongS23}, this field of research has followed a trajectory similar to text-based emotion analysis in NLP. That is, early work largely relied on the use of hand-crafted features such as histograms of oriented gradients \cite{Dalal05}, local binary patterns \cite{Shan05}, or scale-invariant keypoints \cite{Lowe04}  which served as the input to traditional discriminative classifiers such as support vector machines. 
In the past decade, this approach has been replaced by  the use of deep learning and, in particular, convolutional neural networks (CNN) \cite{Goodfellow15,Mollahosseini19,ZhouF20}, while most recently vision transformers have attracted a lot of attention \cite{Huang21,Ma23,XueF23}.

Our work complements these lines of research by providing a method 
that allows existing models to embed the emotional loading of an arbitrary piece of content (in this study, words, texts or images) in a \textit{common} emotion embedding space. This broadens the range of emotional nuances previous models can capture. Importantly, our method learns a representation not for a specific piece of content itself, but the emotion attached to it. This differs from earlier work aiming to increase the affective load of, e.g., word embeddings (see below).

\subsection{Emotion Embeddings}

Several researchers have already introduced the term \enquote{emotion embeddings} (or similar phrasings) to characterize their work, yet either use the term in a different way or tackle a different problem compared to our approach.

Wang et al.\ \cite{Wang20tit}, for instance, present a method for increasing the emotional content of word embeddings based on re-ordering vectors according to the similarity in their emotion values, referring to the result as \enquote{emotional embeddings}. Similarly, Xu et al.\ \cite{Xu18wassa} learn word embeddings that are particularly rich in affective information by sharing an embedding layer between models for different emotion-related tasks. They refer to these embeddings as \enquote{generalized emotion representation}. 
Different from our work, these two studies primarily learn to represent \textit{words} (with a focus on their affective meaning though), not emotions themselves. They are thus in line with previous research aiming to increase the affective load of word embeddings \cite{Faruqui15naacl,TangD16,Yu17emnlp,Khosla18,Shah22coling}.

Shaptala et al.\ \cite{Shantala18} improve a dialogue system by augmenting its training data with emotion predictions from a separate system. Predicted emotion labels are fed into the dialogue model using a representation (\enquote{emotion embeddings}) learned in a supervised fashion with the remainder of the model parameters. These embeddings are specific to their architecture and training dataset, they do not generalize to other label formats. 
Gaonkar et al.\ \cite{Gaonkar20acl} as well as Wang et al.\ \cite{Wang21acl} learn vector representations for emotion classes from annotated text datasets to explicitly model their semantics and inter-relatedness. Yet again, these emotion embeddings (the class representations) do not generalize to other datasets and label formats. 
Han et al.\ \cite{Han19taffc} propose a framework for learning a common embedding space as a means of joining information from different modalities in multimodal emotion data. While these embeddings generalize over different media (audio and video), they do not generalize across languages and label formats.

In summary, different from these studies, our understanding of emotion embeddings is not bound to any particular model architecture or dataset, but instead generalizes across domains and label formats, thus allowing to directly compare, say, English language items with BE5 ratings to Mandarin ones with VA ratings (see Table \ref{tab:examples} vs.\ \fig \ref{fig:emotion-space}).

\subsection{Coping with Incompatibility}
In face of the large variety of emotion formats, Felbo et al.\ \cite{Felbo17emnlp} present a transfer learning approach in which they pre-train a model with self-supervision to predict emojis in a large Twitter dataset, thus learning a representation that captures even subtle emotional nuances. Park et al. \cite{Park21} present a method to learn to predict dimensional emotions from categorical annotations, using the VAD scores of category-denoting words as a simple heuristic. Similarly, multi-task learning can be used to fit a model on multiple datasets potentially having different label formats, thus resulting in shared hidden representations  \cite{Tafreshi18,Augenstein18naacl}. While representations learned with these approaches generalize across different label formats, they do not generalize across model architectures or language domains.

Cross-lingual approaches learn a common latent representation for different languages but these representations are often specific to only \textit{one pair of languages} and do not generalize to other label formats \cite{Gao15cl,Abdalla17ijcnlp,Barnes18acl}.  
Similarly, recent work with Multilingual BERT \cite{Devlin19naacl} shows strong performance in cross-lingual zero-shot transfer \cite{Lamprinidis21wassa}, but samples from different languages still end up in different regions of the embedding space \cite{Pires19acl}. These approaches are also specific to a particular model architecture so that they do not naturally carry over to, e.g.,  single-word emotion prediction. 

Multi-modal approaches to emotion analysis exhibit a high degree of similarity to our work, as they learn a common latent representation for several modalities   \cite{Zadeh17,Han19taffc,Poria19acl,Liang22}. However, these representations are typically specific for a single dataset, model architecture, and emotion label format. They are neither meant to generalize further, nor are they stable across multiple executions.

In a survey on text emotion datasets, Bostan et al.\ \cite{Bostan18coling} point out naming inconsistencies between label formats. They build a joint resource that unifies twelve datasets under a common file format and annotation scheme. Annotations were unified based on the  semantic closeness of their class names (e.g., merging \textit{\enquote{happy}} and \enquote{\textit{Joy}}).  This approach is limited by its reliance on \textit{manually} crafted rules which are difficult to formulate, especially for numerical label formats.

Another attempt at taming the heterogeneity of \enquote{emotion languages} is based on \textit{emotion ontologies} or
\textit{emotion mark-up languages}
\cite{SanchezRada16,Burkhardt17}. In this approach, the shared knowledge about emotions is formulated in some knowledge representation or ontology representation language (such as \textsc{RDFS}, \textsc{SKOS}, \textsc{Owl}, etc.) as a common conceptual backbone to which individual emotion annotations in concrete datasets have to be mapped. Yet, not only the ontology itself, but also the mapping conditions for each emotion model, again, have to be specified manually.\footnote{Cf.\ \cite{Baldoni12,Francisco13} for exceptions to this rule, which still suffer from built-in language dependence.}

In contrast, emotion representation mapping (or \enquote{label mapping})  aims at \textit{automatically} learning such conversion schemes between different formats from data (especially from \enquote{double-annotated} samples, such as the first two rows in Table \ref{tab:examples}; \cite{Buechel18coling}). As the name suggests, label mapping operates exclusively on the gold ratings, without actually deriving representations for language items. It can, however, be used as a post-processor, converting the prediction of another model to an alternative label format (used as a baseline in \sec\ref{sec:setup}). Label mapping learns to transform \textit{one} format \textit{into another}, yet without establishing a more general representation, i.e., $n$-to-$n$ mappings, we here propose. 
In a related study, DeBruyne et al.\ \cite{DeBruyne22csl} indeed do learn a common representation for different label formats by applying variational autoencoders to multiple emotion lexicons. However, their method still only operates exclusively on the gold ratings without actually predicting labels for texts or images.

In summary, while there are methods to learn common emotion representations across \textit{either} heterogeneous modalities, languages, genres, label formats, or model architectures, to the best of our knowledge, our proposal is the first to achieve all this simultaneously, in a homogeneous framework model.

\section{Methods}\label{sec:methods}

\begin{figure*}[tb]
	\centering
    \includegraphics{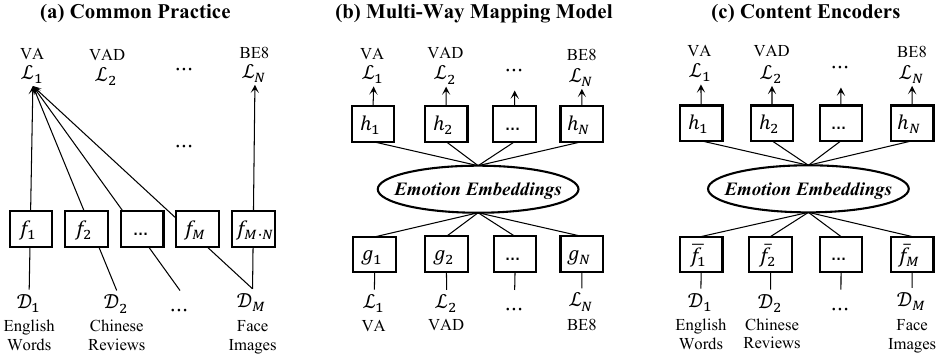}
	\caption{Overview of our methodology, illustrated by several choices of language domains and label formats.
	}
	\label{fig:architecture}
\end{figure*}

The goal of this study is to devise a stable, uniform representation for affective states that can be used for emotion prediction problems across modalities and label formats. It is common practice in emotion analysis---and supervised learning, in general---to build or fine-tune a model specifically for a single dataset. A \textit{dataset} is characterized by a particular combination of domain (its expressional form of modality, genre, or language) and label format (the representational system for emotional meta-data). Considering all existing emotion data, this results in a large number of prediction models, each one covering only a fraction of the problem space and learning \textit{ad hoc} representations that are mostly incompatible to the other models' representations (\fig\ref{fig:architecture}a).  Our method instead learns stable and universal representations (we call \textit{emotion embeddings}) so that two pieces of content that are associated with the similar emotions will receive similar representations, even if these pieces of content come in different modalities and, therefore, require different model architectures for processing. 

We achieve this goal via a two step process: First, we learn a model that can \enquote{translate} between different label formats. Unlike previous work that could only translate between single \textit{pairs} of label formats, our \textit{multi-way mapping model} translates between multiple label formats via a shared intermediate representation which acts as an \enquote{interlingua for emotion} and, at the same time, constitutes the proposed emotion embeddings (\fig\ref{fig:architecture}b). In the second step, we train prediction models, \textit{content encoders}, for specific content types (words, texts, or images) to produce not the final label of a piece of content but its emotion embeddings instead, using the previously built mapping model as a teacher. Emotion labels are then predicted by running a specific content encoder and a label decoder in succession (\fig\ref{fig:architecture}c).

\subsection{Model Architecture and Emotion Inferences}\label{subsec:modarc}

Let  $(X,Y)$ be a dataset with samples $X {\coloneqq} \{x_1, \dots x_n\}$ and labels $Y {\coloneqq} \{y_1, \dots, y_n\}$. Let us  assume that the samples $X$ are drawn from one of $M$ domains $\mathcal{D}_1, \dots, \mathcal{D}_M$ and the labels are drawn from one of $N$ label formats $\mathcal{L}_1, \dots, \mathcal{L}_N$. The common practice for emotion prediction problems is to find a model $f$ that best predicts $Y$ given $X$.   
Our approach instead makes use of a collection of three types of components, \textit{label encoders} $g_1, \dots, g_N$, \textit{label decoders} $h_1, \dots, h_N$, and \textit{content encoders} $\bar{f}_1, \dots, \bar{f}_M$ all connected via a shared intermediate space $\mathbb{R}^d$ housing the emotion embeddings.

\textbf{Label encoders} receive an emotion label and predict its associated emotion embedding. They answer the question what the meaning of a particular label is, following a specific label format, in terms of our general emotion embedding. Label encoders will be implemented as feed-forward networks (FFN; see \S\ref{sec:setup} for details). For each label format under consideration (the choices for this study are depicted in Table \ref{tab:formats}), there is a specific label encoder, and vice versa.

\begin{table}[t!]
    \caption{Emotion Label Formats Considered in this Article}
    \label{tab:formats}
    \centering
    \begin{tabular}{lp{4.5cm}ll}
    \hline
     ID  & Variables & Range & Problem \\
     \hline
      VA   &  Valence, Arousal & $[-1, 1]$ & regress. \\
      VAD &  Valence, Arousal, Dominance & $[-1, 1]$ & regress. \\
      BE5 & Joy, Anger, Sad, Fear, Disgust & $[0, 1]$ & regress. \\
      BE7 & Happy, Anger, Sad, Fear, Disgust, Surprise, Neutral & $\{0, 1\}$ & class. \\
      BE8 & Happiness, Anger, Sadness, Fear, Disgust, Surprise, Neutral, Contempt & $\{0, 1\}$ & class. \\
      \hline
    \end{tabular}
\end{table}

\textbf{Label decoders}, conversely, take an emotion embedding and produce a label in a particular emotion format. Again, for each such format, there is a particular label decoder (and vice versa). Label decoders can be thought of as the output layer in a regular emotion prediction model, converting the top-level emotion representation into the predicted label. Thus, we will alternatively refer to them as \textbf{prediction heads}.
We give these heads a purposefully lean design that consists only of a single linear layer without bias term, followed by an optional activation function depending on the respective label format and its implied learning problem. 

Thus, a head $h$ predicts ratings $\hat{y}$ for an emotion embedding $x\in \mathbb{R}^d$ as $h(x) \coloneqq z(W x)$, where $W$ is a weight matrix and $z$ the activation function. $z$ takes the value of the softmax function for multi-class single-label classification, the sigmoid function for multi-class multi-label classification, or the identity, i.e., no further activation, for regression problems. The reason for this simple head design is to ensure that the affective information is more readily available in the emotion space. 

\textbf{Content encoders}, finally, are most similar to the usual models in emotion prediction problems, only that their output is not a final label but rather the predicted emotion embedding for a particular piece of content. As with their baseline counterparts, their architecture needs to account for the type of content they are supposed to handle. We will use FFNs for word-type input, transformers for text-type input, and CNNs for image-type input. In addition, datasets can also differ in more subtle ways, i.e., the language or genre of its sample. These differences do not necessarily demand a different model architecture, but still different model parameters for good performance. Hence, each content encoder is specific to a particular domain. Thus, our approach imposes no architectural constraints on what can serve as a content encoder, other than dense, vector-valued output needs to be produced. Virtually any deep learning architecture---including those not yet invented---can, in principle, serve as a content encoder within our framework.

Label encoders and label decoders together constitute our proposed multi-way mapping model. By training a collection of these components in such a way that they all share a common intermediate representation, encoders and decoders can be freely combined to translate between any pair of label formats. Formally, to translate a label $y_i$ in format $\mathcal{L}_i$ to its predicted label $\hat{y}_j$ in format $\mathcal{L}_j$ one only has to apply the respective components in succession: $\hat{y}_j = h_j(g_i(y_i))$. Similarly, after a content encoder $\bar{f}_k$ for a domain $\mathcal{D}_k$ is trained, predictions for label format $\mathcal{L}_l$ can be derived as $\hat{y}_l = h_l(\bar{f}_k)$. We provide more details of the devised training process in the following two subsections.

\subsection{Training the Multi-Way Mapping Model}
\label{sec:mapping-training}

\begin{algorithm*}
	\caption{Training the Multi-Way Mapping Model}
	\label{alg:pretraining}
	\begin{algorithmic}[1]
\STATE $(Y_{1,1}, Y_{1,2}), (Y_{2,1}, Y_{2,2}), \dots (Y_{n,1}, Y_{n,2}) \leftarrow$ Mapping datasets used for training 
\STATE $ (\mathcal{C}_{1,1}, \mathcal{C}_{1,2}), (\mathcal{C}_{2,1}, \mathcal{C}_{2,2}) \dots, (\mathcal{C}_{n,1}, \mathcal{C}_{n,2}) \leftarrow$  loss criteria for mapping datasets
\STATE $ (\alpha_{1,1},\alpha_{1,2}), (\alpha_{2,1},\alpha_{2,2}), \dots (\alpha_{n,1},\alpha_{n,2}) \leftarrow $ weight terms to balance between loss criteria 
\STATE $(g_{1,1}, h_{1,1}, g_{1,2}, h_{1,2}), (g_{2,1}, h_{2,1}, g_{2,2}, h_{2,2}), \dots, (g_{n,1}, h_{n,1}, g_{n,2}, h_{n,2})\leftarrow$  randomly initialized label encoders and prediction heads \textsuperscript{\dag}
\STATE $\mathcal{K}=\{K_1, K_2, \dots, K_m\} \leftarrow $ Set of equivalence classes of emotion variables, where each equivalence class $K_j$ is a set of row vectors from weight matrices of different prediction heads that refer to equivalent emotion variables (see Table \ref{tab:variable-equivalence}).

\STATE $n_\mathrm{steps} \leftarrow$ total number of training steps
\FORALL{$k$ in $1, \dots, n_\mathrm{steps}$}
\STATE $i \leftarrow$ randomly sample index of one of the mapping datasets $\in \{1,\dots, n\}$ \label{algline:start-sampling}
\STATE $(y_1, y_2) \leftarrow$  randomly sample a batch s.t. $y_1 \subset Y_{i,1}$ and $y_{2} \subset Y_{i,2}$ with identical indices \label{algline:stop-sampling}
\STATE $(e_1, e_2) \leftarrow (g_{i,1}(y_1), g_{i,2}(y_2))$ \label{algline:encoding}
\STATE $\hat{y}_{1,1} \leftarrow h_{i,1}(e_1)$ \label{algline:start-decoding}
\STATE $\hat{y}_{1,2} \leftarrow h_{i,2}(e_1)$
\STATE $\hat{y}_{2,1} \leftarrow h_{i,1}(e_2)$
\STATE $\hat{y}_{2,2} \leftarrow h_{i,2}(e_2)$  \label{algline:stop-decoding}
\STATE $L_\mathrm{map} \leftarrow \mathcal{C}_{i,1}(y_1,\hat{y}_{2,1}) + \mathcal{C}_{i,2}(y_2, \hat{y}_{1,2}) $ 
\STATE $L_\mathrm{auto} \leftarrow \mathcal{C}(y_1,\hat{y}_{1,1}) + \mathcal{C}(y_2, \hat{y}_{2,2})$ \label{algline:auto} 
\STATE $L_\mathrm{sim} \leftarrow \mathcal{C}_\mathrm{sim}(e_1,e_2)$ \label{algline:sim} 
\STATE $L_\mathrm{para} \leftarrow  \sum_{K \in \mathcal{K}} \sum_{\;\;\{u,v\ \vert u,v \in K, u \neq v\}} 1 - \mathrm{cos}(u,v) $
\STATE $L_\mathrm{total} \leftarrow L_\mathrm{map} + L_\mathrm{auto} + L_\mathrm{sim} + L_\mathrm{para}$ \label{algline:start-update}
\STATE compute $\nabla L_\mathrm{total}$ and update weights \label{algline:stop-update}
\ENDFOR
\par\noindent\rule{.2\columnwidth}{0.4pt}
\par\noindent {
\textsuperscript{\dag} If two sets of labels $Y_{a,b}, Y_{c,d}$ follow the same label format, then they use the same label encoders and prediction heads, i.e, $g_{a,b} = g_{c,d}$ and  $h_{a,b} = h_{c,d}$. The same is true for loss criteria $\mathcal{C}$ and weight terms $\alpha$.\\ 
}
\end{algorithmic}

\end{algorithm*}

It should be noted that the emotion embeddings, which are simply elements of a Euclidean vector space, carry no intrinsic affective meaning by themselves. Rather, explicit meaning is given to them by the label encoders $g_1, \dots, g_N$ and the label decoders $h_1, \dots, h_N$, because they prescribe which affective state a region of the representation space corresponds to. The problem that our proposed training scheme solves is that all encoders and decoders need to be consistent in their meaning prescriptions.

But what  does \enquote{consistency} mean when emotion labels follow different representational systems? We argue that an obvious case of such consistency conditions is found in datasets for emotion label mapping (see \sec \ref{sec:related}). A label mapping dataset consists of two sets of  labels following different formats $Y_1 {:=} \{y_{1,1},  y_{1,2},\dots y_{1,n}\}$ and $Y_{2} {:=} \{y_{2,1}, y_{2,2}, \dots y_{2,n}\}$, respectively.
Typically, they are constructed by matching instances from independent annotation studies (e.g., the first two rows in Table \ref{tab:examples}). Thus, we can think of the two sets of labels as \enquote{translational equivalents}, i.e., differently formatted emotion ratings, possibly capturing different affective nuances, yet still describing the same underlying emotional state. 

The intuition behind our training scheme is to \enquote{fuse} multiple mapping models by forcing them to produce the same intermediate representation for both mapping directions.  This results in a multi-way mapping model with a shared representation layer in the middle, the common emotion space (\fig\ref{fig:architecture}\textit{b}).

\begin{figure}[b!]
    \centering
    \includegraphics{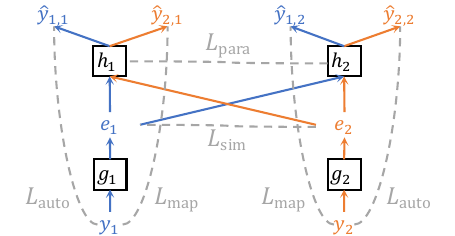}
    \caption{Training the Multi-Way Mapping Model. }
    \label{fig:head-training}
\end{figure}

Formally, let $(Y_1, Y_2)$ be a mapping dataset with a sample $(y_1, y_2)$ following label formats $\mathcal{L}_1, \mathcal{L}_2$ (subscripts refer to label format). 
Our goal is to align the intermediate representations, $e_1, e_2 \coloneqq g_1(y_1)$, $g_2(y_2)$  while also deriving accurate mapping predictions. Therefore, we propose the following three training objectives (see  \fig \ref{fig:head-training}):

The first training objective is the  \textit{mapping loss} $L_\mathrm{map}$ where we compare true vs. predicted labels.\vspace*{-10pt}

\begin{equation}
     L_\mathrm{map} \coloneqq \alpha_1\mathcal{C}_1[y_1, h_1(g_2(y_2))] + \alpha_2\mathcal{C}_2[y_2 , h_2(g_1(y_1))]
\end{equation}

\noindent The two summands represent the two mapping directions, assigning either of the two labels as the source, the other as the target format. $\mathcal{C}_i$ denotes the loss criterion for label format $\mathcal{L}_i$ and $\alpha_i$ is the format-specific weight term which is meant to re-balance loss criteria that otherwise tend to produce outputs of vastly different magnitude. As loss criteria, we use cross-entropy loss for multi-class single-label problems, binary cross-entropy loss for multi-class multi-label problems, and Mean-Squared-Error loss for regression problems.

Next, the  \textit{autoencoder loss}, $L_\mathrm{auto}$, captures how well the model can reconstruct the original input label from the hidden emotion representation. It is meant to supplement the mapping loss: 
\begin{equation}
     L_\mathrm{auto} \coloneqq \alpha_1\mathcal{C}_1[y_1 , h_1(g_1(y_1))] + \alpha_2 \mathcal{C}_2[y_2, (h_2(g_2(y_2))] 
\end{equation}

Lastly, the \textit{similarity loss}, $L_\mathrm{sim}$, directly assesses whether both input label formats end up with a similar intermediate representation:
\begin{equation}
     L_\mathrm{sim} \coloneqq \mathcal{C}_\mathrm{sim}[g_1(y_1), g_2(y_2)]
\end{equation}
where $\mathcal{C}_\mathrm{sim}$ denotes Mean-Squared-Error loss.

Besides these \textit{local} loss terms which only take into account encoders and decoders used in processing a particular instance (pair of labels), we also introduce a parameter sharing loss $L_\mathrm{para}$ that \textit{globally} takes into account \textit{all} label decoders of the multi-way mapping model, implementing \textit{soft parameter sharing} between them, a well-known technique for multi-task learning \cite{Ruder17}. Unlike \textit{hard} parameter sharing where a model uses the same (sub)set of parameters for different tasks, in soft parameter sharing, model parameters between tasks do not need to be strictly identical. Rather, a loss function is applied to incentivize that parameter sets (mostly) agree with each other. 

We use soft parameter sharing between label encoders as a way to ensure that equivalent emotion variables in different label formats have similar heads (e.g., Joy in BE5 and Happiness in BE8), while also giving the model some flexibility to adjust to the particularities of the individual datasets (we validate this approach in \S\ref{sec:visualization}). In particular, note that the prediction heads may end up predicting logits (the vector-matrix product $Wx$ before applying the activation function $z$) of different value ranges. E.g., when comparing Joy in BE5, which is a regression format, so the output of the Joy head needs to be a specific number, to Happiness in BE8, which is a multi-class format, so the output needs to be much larger than the logits for the remaining variables. To allow parameters of equivalent heads to be \enquote{similar}, yet produce logits of different magnitude, we turn to a cosine-based loss function so that heads (interpreted as weight vectors) are only penalized for \enquote{pointing to different directions}, but not for differences in vector length.

\begin{table}[b!]
    \caption{Equivalence Classes of Emotion Variables for Soft Parameter Sharing}
    \label{tab:variable-equivalence}
    \centering
    \begin{tabular}{l}
    \hline
    VA:Valence---VAD:Valence \\
    VA:Arousal---VAD:Arousal \\
    BE5:Joy---BE7:Happy---BE8:Happiness \\
    BE5:Anger---BE7:Anger---BE8:Anger \\
    BE5:Sadness---BE7:Sad---BE8:Sadness \\
    BE5:Fear---BE7:Fear---BE8:Fear \\
    BE5:Disgust---BE7:Disgust---BE8:Disgust \\
    BE7:Surprise---BE8:Surprise \\
    BE7:Neutral---BE8:Neutral \\
    \hline
    \end{tabular}
\end{table}

More formally, we define a set of equivalence classes of  emotion variables, i.e., emotion variables from different label formats for which we stipulate that they have identical affective meaning (see Table \ref{tab:variable-equivalence}). 
We then sum up the cosine loss for each combination of variables (respectively, prediction heads) in each of those equivalence classes:

\begin{equation}
    L_\mathrm{para} \coloneqq \sum_{K \in \mathcal{K}} \sum_{\;\;\{u,v\ \vert u,v \in K, u \neq v\}} 1 - \mathrm{cos}(u,v)
\end{equation}

\noindent where $\mathcal{K}$ denotes the set of all equivalence classes, $K$ denotes either of those equivalence classes, and $u,v$ denote equivalent weight vectors (heads of the equivalent variables) in the emotion embedding space $E$.

The total loss is computed in a batch-wise fashion detailed in  Algorithm \ref{alg:pretraining}: We simultaneously train all label encoders and decoders on several mapping datasets using our four objective functions. 
Our training loop starts by first sampling one of the mapping datasets and then a batch from the chosen dataset (lines \ref{algline:start-sampling}--\ref{algline:stop-sampling}).  To compute the loss efficiently, we first cache the encoded representations of both label formats (line \ref{algline:encoding}) before applying all relevant label encoders (lines \ref{algline:start-decoding}--\ref{algline:stop-decoding}). The individual losses are then summed up and the model is updated accordingly (\ref{algline:start-update}--\ref{algline:stop-update}).

\subsection{Training the Content Encoders}
\label{sec:deployment}

After training the multi-way mapping model, we have label encoders, fitted to map different kinds of emotion ratings into the emotion embedding space, and label decoders, fitted to reverse this process. What is still missing are components to embed pieces of content into the emotion space as well. These content encoders are constructed by taking an existing model architecture for a specific content type (e.g., a CNN for images) and replacing its output layer by a linear layer of output size $d$, thus matching the dimensionality of the emotion embedding space. The rest of the model may remain unchanged.

Content encoders are trained in a supervised fashion on a given dataset by combining them with a suitable label decoder (i.e., a label decoder that matches the label format of the dataset; see above) using common gradient descent-based optimization. The parameters of the label decoder are kept constant during this process, thus forcing the content encoder to produce meaningful output representations, i.e.,  emotion embeddings that are consistent with the label decoders' \enquote{understanding} of the emotion space. 

\begin{figure}[b!]
    \centering
    \includegraphics{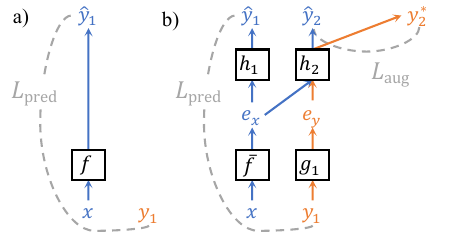}
    \caption{Fitting a domain-specific emotion prediction model:  Baseline (a) vs. our method (b).}
    \label{fig:content-encoder-training}
\end{figure}

To make sure that the content encoder inter-operates well with the remaining label decoders, too, we add additional supervision via a newly proposed data augmentation technique that we call \textbf{emotion label augmentation}. This method generates an additional label, following another format, on the fly using the previously trained multi-way mapping model. In other words, the multi-way mapping model is used as a teacher, providing additional supervision to the content encoder, the student (see \fig\ref{fig:content-encoder-training}).

Formally, let $(x,y_1)$ be an instance from a dataset following format $\mathcal{L}_1$, let $\bar{f}$ be a content encoder suitable for the domain of the dataset, let, $h_1,h_2$ be label decoders for $\mathcal{L}_1, \mathcal{L}_2$, respectively, and let $g_1$ be a label encoder for $\mathcal{L}_1$. Then the regular prediction loss is given by

\begin{equation}
  L_\mathrm{pred} \coloneqq \mathcal{C}_1[y_1,\hat{y}_1]  
\end{equation}

\noindent where $\hat{y}_1 \coloneqq h_1(\bar{f}(x))$. Emotion label augmentation synthesizes an additional label $y^*_2 = h_2(g_1(y_1))$ which can then be used to calculate a complementary data augmentation loss

\begin{equation}
L_\mathrm{aug} \coloneqq \mathcal{C}_2[y^*_2, \hat{y}_2]     
\end{equation}

\noindent where $\hat{y}_2 \coloneqq h_2(\bar{f}(x))$. 
The total loss $L_\mathrm{pred}+L_\mathrm{aug}$ is then optimized in batch-wise training. 

The result of this process is that the collective model can not only reliably predict ratings according to format $\mathcal{L}_1$ but also $\mathcal{L}_2$, even in the absence gold data following this format. Thus, it grants our collective model the ability for zero-shot learning.

\section{Experiments}\label{sec:setup}

\begin{figure*}
    \centering
    \includegraphics{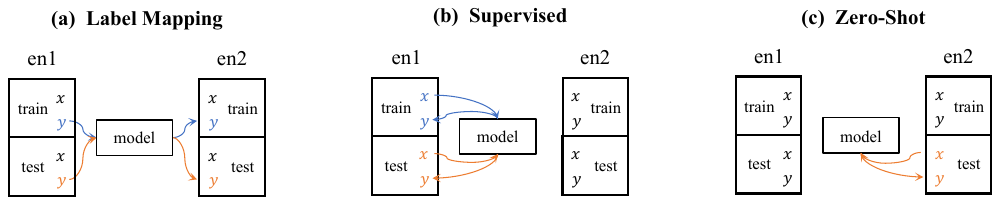}
    \caption{The three scenarios of our experimental setup exemplified with the word datasets en1 (in VAD format) and en2 (in BE5 format). Blue color indicates data usage for training, orange for testing. Dev sets are omitted for simplicity. }
    \label{fig:experiments}
\end{figure*}

\textbf{Overall Approach.}\
Our experiments follow a two-fold structure akin to that of our proposed training process. In the first phase of our experiments, we train the multi-way mapping (consisting of label encoders and decoders) and evaluate it against existing models for emotion label mapping. This establishes the emotion embeddings space that \textit{all} components in \textit{all} of our experimental setups share. In the second phase, we train the content encoders (using the multi-way mapping model as a teacher) and evaluate them against existing models in emotion analysis. Here, we distinguish two settings, first, a \textit{supervised} setting where training and test data stem from the same ecosystem, and, second, a \textit{zero-shot} setting where the label format of the test data differs from that of the training data (see \fig\ref{fig:experiments}).

With this setup, we aim to demonstrate three major claims underlying our work: first, our approach enables predictions for diverse domains and label formats via a \textit{shared} intermediate representation of emotions; second, our approach enables predictions for label formats it was \textit{not trained on} and where training data is \textit{potentially unavailable}; and third, prediction quality remains competitive to common practice, i.e., our approach yields the aforementioned benefits \textit{without performance penalties}. In other words, our goal is to establish a new state of the art (SOTA) in model flexibility and interoperability, rather than sheer quantitative performance. Furthermore, we will show that the performance of our approach directly depends on the model types used as content encoders. We thus conjecture that the performance of our approach will improve \textit{virtually by itself} as the SOTA throughout the affective computing community advances.

\begin{table}[b!]
    \caption{Datasets used in experiments}
    \label{tab:data}
    \centering
    \begin{tabular}{lllrr}
    \hline
    \textbf{ID} & \textbf{Domain} & \textbf{Format} &  \textbf{Size} & \textbf{Source} \\
    \hline
    \hline
    en1     & English words  &  VAD &   1,034    &   \cite{Bradley99anew} \\
    en2     & English words  &  BE5 &   1,034    &   \cite{Stevenson07} \\
    es1     & Spanish words  &  VA  &  14,031    &   \cite{Stadthagen17} \\
    es2     & Spanish words  &  BE5 &  10,491    &   \cite{Stadthagen18} \\
    de1     & German words   &  VA  &   2,902    &   \cite{Vo09} \\
    de2     & German words   &  BE5 &   1,958    &   \cite{Briesemeister11} \\
    pl1     & Polish words   &  VA  &   2,902    &   \cite{Riegel15}            \\
    pl2     & Polish words   &  BE5 &   2,902    &   \cite{Wierzba15}            \\
    tr1     & Turkish words  &  VA  &   2,029    &   \cite{Kapucu18}            \\
    tr2     & Turkish words  &  BE5 &   2,029    &   \cite{Kapucu18}            \\
    \hline
    AffT    & English news headlines & BE5          & 1,250 &  \cite{Strapparava07}  \\
    EmoB    & mixed English texts    & VAD & 10,062 & \cite{Buechel17eacl} \\
    \hline
    ferBE  & facial images    &  BE7 & 35,887     & \cite{Goodfellow15} \\
    ferVAD & facial images    &  VAD & 35,887     & \cite{ZhouF20} \\
    afnBE  & facial images    &  VA  &  450,000   & \cite{Mollahosseini19}\\
    afnVA  & facial images    &  BE8 &   450,000  & \cite{Mollahosseini19}   \\
    \hline
    \end{tabular}
\end{table}

\textbf{Data and Metrics. } The experiments will be conducted on a total of 16 datasets from three different modalities, i.e., containing words, texts, and images. Table \ref{tab:data} lists the employed datasets; illustrative examples are given in Table \ref{tab:examples}. Note that word and text datasets cover different natural languages. The datasets are partitioned into fixed train-dev-test splits with ratios ranging between 8-1-1 and 3-1-1; smaller datasets received larger dev and test shares. Details on data handling are documented as part of our codebase (see Abstract for availability information).

We will use different evaluation metrics depending on the label format of the respective dataset and its corresponding prediction problem (i.e., classification vs. regression; see Tables \ref{tab:data} and \ref{tab:formats}). In more detail, for regression datasets, model output is evaluated in terms of Pearson Correlation $r$: 

\begin{equation*}
	r_{x,y} \coloneqq \frac{\sum_{i=1}^n (x_i-\bar{x})(y_i-\bar{y})}{\sqrt{\sum_{i=1}^n(x_i-\bar{x})^2} \; \sqrt{\sum_{i=1}^n(y_i-\bar{y})^2}}
\end{equation*}

\noindent where $x = x_1, x_2, \dots, x_n$, $y = y_1, y_2, \dots, y_n$ are real-valued number sequences and $\bar{x}$, $\bar{y}$ are their respective means. In contrast, classification predictions will be evaluated in terms of accuracy, i.e., $\frac{\text{No. true predictions}}{\text{No. total predictions}}$.

\textbf{Evaluating the Multi-Way Mapping Model.} We trained the multi-way mapping model on four label mapping datasets in parallel. These datasets were created by combining data from Table \ref{tab:data} which had overlapping samples: en1~$\cap$~en2, en1*~$\cap$~en2, ferVAD~$\cap$~ferBE, and afnVA~$\cap$~afnBE, where en1* refers to a modified version of en1 without dominance ratings. This collection of datasets allowed us to train label decoders for a total of five label formats listed in Table \ref{tab:formats}. As detailed in \S\ref{sec:mapping-training}, label mapping operates exclusively on the \textit{labels}, not the samples, of a dataset (see \fig \ref{fig:experiments}a).

Label encoders follow the design of the two-hidden-layer FFN from Buechel and Hahn \cite{Buechel18coling}, which to the best of our knowledge still represents the current SOTA in emotion label mapping. The same model type (stand-alone, not as a label encoder within our framework) was also used as a baseline model for experimental comparison. For each of the mapping datasets, we trained two baseline models (one for each mapping direction) for a total of individual 8 baseline models. We compared these ten models against our multi-way mapping models which consists of five label encoders and five label decoders, but has a single shared intermediate layer (housing the emotion embeddings). The dimensionality of this intermediate layer was set to 100.

\textbf{Evaluating the Content Encoders.}
The main idea for this evaluation task is to take a popular architecture for each of the three modalities under consideration (words, texts, and images) and compare its performance as a stand-alone model (following common practice) with the performance achieved when using it as a content encoder within our framework instead. Again, we aim at demonstrating that our framework \textit{increases} interoperability and flexibility of models and datasets while \textit{retaining} prediction quality. 

For our two evaluation scenarios (see below), datasets need to be grouped into pairs where both members of each pair agree in the domain of their data but employ distinct label formats. The groupings are indicated in Table \ref{tab:data}, where word and image datasets are grouped by their ID prefix (e.g., en1 and en2 form a group) and among the text datasets \textsc{EmoB[ank]} and \textsc{AffT[ext]} form a group on their own. 
We use min-max scaling to normalize value ranges of the labels across regression datasets: for VA(D) we choose the interval $[-1, 1]$ and for BE5 the interval $[0, 1]$, reflecting their respective bipolar (VAD) and unipolar (BE5) nature.

The actual evaluation proceeds according to either of the two scenarios: In the \textit{supervised} scenario, both in the baseline and the experimental condition, the respective model was trained on the train set and then tested on the test set (\fig\ref{fig:experiments}b). In the experimental condition, the model consists of the respective content encoder combined with the label decoder (trained during the label mapping experiments) of the respective label formats with frozen weights. Additionally, we employ the label decoder of the format of the \textit{other} dataset of the pair to create synthetic labels for additional supervision (the emotion label augmentation technique presented in \S\ref{sec:deployment}). In the baseline condition, the base model of the content encoder is used in a stand-alone fashion.

In the \textit{zero-shot} scenario, no training occurs. For each dataset, we take the model trained in the supervised scenario for the \textit{other} dataset within the pair and evaluate it on its test set straight away (see \fig\ref{fig:experiments}c). The test data follow a different label format compared to the train data, hence our use of the term \enquote{zero-shot (learning)}. In the experimental conditions (using our framework) this is possible because the required label decoder has already been trained during the label mapping experiments. Also the respective content encoder has been tuned to interoperate with said label decoder through the use of label augmentation during the supervised scenario (see above). The baseline model \textit{per se}, however, is simply unable to predict the label format of the test data. To still be able to offer a quantitative comparison, we resort to the label mapping baseline model from scenario (a) that translates the base model's output into the desired format. We emphasize that this is a very strong baseline due to the high accuracy of the label mapping approach, in general \cite{Buechel18coling}. In this scenario, the supremacy of our approach lies in its independence from these (possibly unavailable) external post-processors. 

In terms of content encoder architectures, we used the FFN developed by Buechel and Hahn \cite{Buechel18naacl} for the word datasets. This model predicts emotion ratings based on pre-trained embedding vectors (taken from \cite{Grave18lrec}). For text datasets, we chose the $\textsc{Bert}_\mathrm{base}$ transformer model by Dev\-lin et al.\ \cite{Devlin19naacl} using the implementation and pre-trained weights by Wolf et al.\ \cite{Wolf20emnlp}. For the image datasets, we used a \textsc{ResNet50} model \cite{He16} pretrained on \textsc{ImageNet} \cite{Deng09}.

Word, text, and image models use identical hyperparameter settings whether or not they are used as stand-alone baseline models or as content encoders within our framework. For the word model, we relied on the original settings, whereas for the text and image models hyperparameter setting were tuned manually for the baseline model and then patched into the respective content encoder. This approach helped reduce manual work for experimentation, but slightly favors the baseline models.

\section{Results}
\label{sec:results}

Our main experimental results are summarized in Tables \ref{tab:results-mapping} and \ref{tab:content-results}. For conciseness, performance scores are averaged over all target variables for regression datasets. We emphasize that all performance figures relating to our proposed framework were achieved based on the same emotion embedding space.

\begin{table}[b!]
    \caption{Results of Label Mapping Experiments}
    \label{tab:results-mapping}
    \centering
    \footnotesize
    \begin{tabular}{|lllrr|}
\hline
Source & Target & Metric &   Baseline &  Ours \\
\hline\hline
en1 & en2       & $r$ &      .89 &       .88 \\
en2 & en1    & $r$     &      .87 &       .87 \\
en1* & en2    & $r$  &      .88 &       .86 \\
en2 & en1*    & $r$   &      .88 &       .88 \\
ferVAD & ferBE   & Acc         &      .66 &       .66 \\
ferBE & ferVAD  & $r$          &      .79 &       .79 \\
afnVA & afnBE    & Acc &      .82 &       .81 \\
afnBE & afnVA   & $r$ &      .93 &       .93 \\
\hline\hline
\multicolumn{3}{|l}{\textbf{Mean}}           &      .84 &       .84 \\
\multicolumn{3}{|l}{\textbf{Joint Embedding Space}} & \XSolidBrush  & \Checkmark\\
\multicolumn{3}{|l}{\textbf{Additional Mapping Directions}} & \XSolidBrush  & \Checkmark\\
\hline
\end{tabular}
\end{table}

\begin{table*}[t!]
    \caption{Results of Content Encoder Experiments}
    \label{tab:content-results}
    \centering
    \begin{tabular}{|l l l | c c | c c |}
    \hline
    &&&  \multicolumn{2}{c|}{\textbf{Baseline}} & \multicolumn{2}{c|}{\textbf{Our Method}}  \\
    Input & Dataset & Metric   & Supervised & Zero-Shot & Supervised & Zero-Shot \\
    \hline\hline
    \multirow{11}{*}{Word (FFN)} & en1 & $r$          & .82 & .79 & .83 & .81 \\
                          & en2 & $r$          & .90 & .84 & .90 & .84 \\
                          & es1 & $r$          & .82 & .72 & .82 & .73 \\  
                          & es2 & $r$          & .79 & .79 & .78 & .77 \\                   
                          & de1 & $r$          & .82 & .68 & .82 & .70 \\                   
                          & de2 & $r$          & .77 & .66 & .77 & .63 \\                   
                          & pl1 & $r$          & .79 & .81 & .79 & .80 \\                   
                          & pl2 & $r$          & .81 & .79 & .80 & .78 \\                   
                          & tr1 & $r$          & .73 & .71 & .73 & .72 \\                   
                          & tr2 & $r$          & .82 & .73 & .82 & .71 \\  
                          \cline{2-7}
                          & Mean  &  & .81 & .75 & .81 & .75 \\ 
    \hline
    \multirow{3}{*}{Text (BERT)} & EmoB       & $r$ & .63 & .37 & .63 & .40 \\  
                          & AffT       & $r$ & .75 & .57 & .76 & .57 \\ 
                          \cline{2-7}
                          & Mean  &    & .69 & .47 & .69 & .48 \\ 
    \hline
    \multirow{5}{*}{Image (ResNet)} & ferBE & Acc    & .70 & .54 & .66 & .63 \\
                           & ferVAD & $r$            & .82 & .78 & .87 & .81 \\
                           & afnBE & Acc          & .51 & .39 & .49 & .46 \\
                           & afnVA & $r$          & .56 & .34 & .58 & .58 \\ 
                           \cline{2-7}
                           & Mean  &     & .65 & .51 & .65 & .62 \\
    \hline
    \hline
    \multicolumn{3}{|l|}{\textbf{Joint Embedding Space}} & \multicolumn{2}{c|}{\XSolidBrush} & \multicolumn{2}{c|}{\Checkmark} \\
    \multicolumn{3}{|l|}{\textbf{Stable Representations}} & \multicolumn{2}{c|}{\XSolidBrush} & \multicolumn{2}{c|}{\Checkmark} \\
    \multicolumn{3}{|l|}{\textbf{Built-In Zero-Shot Capacity}} & \multicolumn{2}{c|}{\XSolidBrush} & \multicolumn{2}{c|}{\Checkmark} \\
    \hline
    \end{tabular}
\end{table*} 
 
As Table \ref{tab:results-mapping} reveals, results of the label mapping experiments remain virtually unchanged between the individual, stand-alone baseline models and our multi-way mapping model, with only minor losses of performance on some datasets. Yet, the advantage of our proposed approach is that all mapping directions are served using a single intermediate representation which allows us to train further content encoders using the same emotion embeddings space (see below). As a positive side effect, the multi-way mapping model can also serve mapping directions for which no training data are available---by combining the respective encoders and decoders (e.g., mapping from BE5 to BE8).

Regarding the results of the content encoder experiments, a similar pattern emerges from Table \ref{tab:content-results}. In the supervised scenario, the performance again remains virtually unchanged compared to the baseline across word, text, and image datasets. The same is true in the zero-shot scenario with the exception of the image datasets. There we see a massive performance advantage of our approach. Upon closer inspection, we conjecture that on the pairs of image datasets the connection between the alternative label formats is weaker than on those of word and text datasets (e.g.,  observe that label mapping baseline performance from ferVAD to ferBE (Table \ref{tab:results-mapping}) is actually \textit{worse} than the baseline-supervised performance (Table \ref{tab:content-results}) which is very uncommon \cite{Buechel18coling}). This seems to massively disturb the baseline zero-shot approach, whereas our approach appears more robust. 

Besides these minor quantitative gains, the major upside of our technique is that all experiments (across all datasets, modalities, and model types) were based on the \textit{same} set of stable, intermediate emotion representations---which may form the basis for entirely new applications (see \S\ref{sec:emotion-retrieval})---and that prediction of alternative label formats (the zero-shot scenario) does not require external mapping components.
Such components may be unavailable for specific formats due to the lack of training data where our encoder-decoder framework may still succeed in producing the respective outputs (\enquote{Additional Mapping Directions} in Table \ref{tab:results-mapping}). In summary,  the presented experiments indicate that our approach reaches its goal of increasing model interoperability and flexibility while retaining original performance levels.

\section{Analysis}\label{sec:analysis}

This section aims at further elucidating the inner workings and future potential of our proposed encoder-decoder architecture and the emotion embeddings it produces.

\label{sec:analyses}

\subsection{Ablation Study}
While we have shown above that our integrated interoperable approach yields mostly on-a-par performance compared with previous isolated non-interoperable approaches, it is not exactly clear how this result was achieved. The following experiment seeks to generate further insights into the importance of the individual proposed techniques.

To get closer to this goal, we conducted three ablation variants of the above experiments. First, in the supervised setup, we evaluated our proposed model \textit{without} applying emotion label augmentation. So the difference between the baseline and the experimental condition is really only the addition of the pre-trained label decoder. Second, also in the supervised scenario, we trained our model \textit{not} with emotion label augmentation, but instead replaced it by \textit{multi-task learning}, training it on both members of a dataset pair in parallel. Third, we evaluated the model trained \textit{without} label augmentation in the zero-shot scenario. Table \ref{tab:ablation} summarizes the results, in terms of average performance over word, text, and image datasets, respectively.

\begin{table}[b!]
    \caption{Ablation Results}
    \label{tab:ablation}
    \centering
    \begin{tabular}{l c c c }
    \hline
    & \textbf{Word} & \textbf{Text} & \textbf{Image} \\
    \hline\hline
    Supervised & .81 & .69 & .65  \\
    --label augmentation & .81 & .68 & .63 \\
    +multi-task learning & .82 & .68 & .59 \\
    \hline
    Zero-Shot             & .75 & .48 & .62 \\
    --label augmentation & .67 & .37 & .37 \\
    \hline
    \end{tabular}
\end{table}

We find that applying (no) label augmentation has only a minor effect on performance in the supervised scenario. Hence, decomposing the base model into a content encoder and a label decoder has only little immediate effect on prediction quality. In the zero-shot scenario, however, performance worsens dramatically without this technique. We can, thus, deduce that label augmentation indeed serves its purpose of increasing the ability of our framework to generate predictions for \textit{all} considered label formats (i.e., increase the interoperability of a content encoder with all label encoders). 

The model variant that uses multi-task learning rather than label augmentation improves on some but deteriorates on other datasets for an overall negative effect. This indicates that the synthetic data introduced by label augmentation work better than actual gold data from a different dataset (as is done with the multi-task approach). Closer inspection suggests that, again, the effect may be mediated by the robustness of the relationship between differently-formatted labels in the dataset pairs, e.g., the noisy relationship between VAD labels in ferVAD and BE7 labels in ferBE seem to severly reduce the performance of the multi-task approach.
In contrast, the label augmentation technique is not so much affected, because the generated synthetic labels are based on the more robust training of the multi-way mapping model. Most importantly, label augmentation is superior to multi-task learning in that it does not rely on additional gold data, a requirement that is not always easy to fulfill.

\subsection{Visualization of the Common Emotion Space} 
\label{sec:visualization}

\begin{figure}[b!]
    \centering
    \includegraphics[width=.49\textwidth]{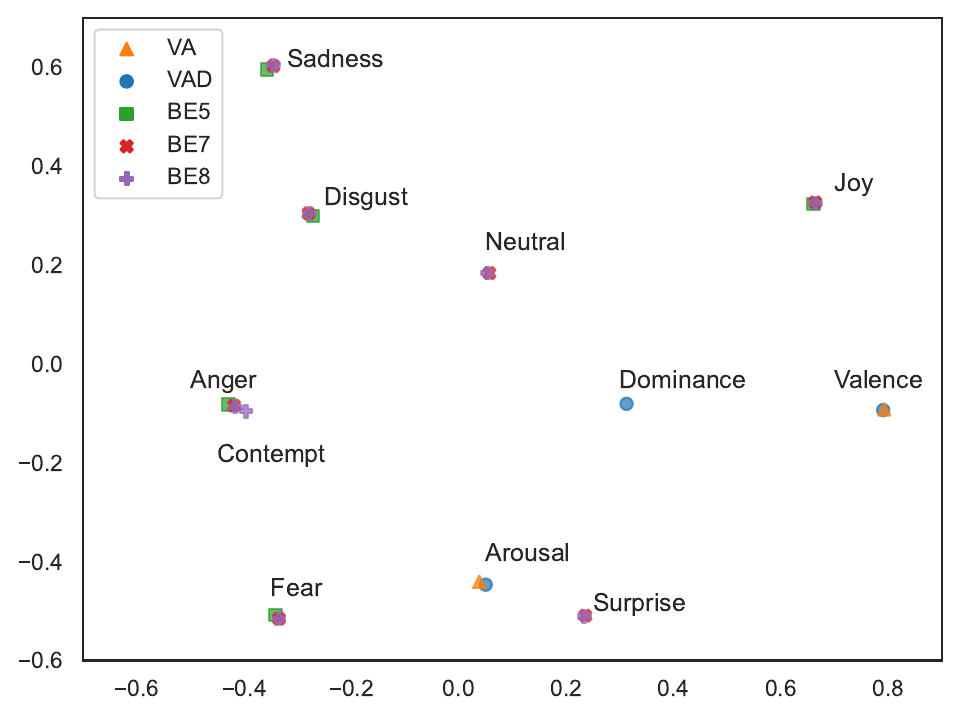}
    \caption{Position of emotion variables in PCA space. Marker type indicates which label format an emotion variable belongs to. }
    \label{fig:variables}
\end{figure}

Next, we visually examine the structure of the shared emotion space underlying the above results. Recall from \S\ref{subsec:modarc} that each label decoder consists of a single weight matrix $W$ of shape $(\text{number of output emotion variables})\times(\text{dimensionality of emotion space})$. Each row vector of $W$ is thus associated with a single emotion variable and, when interpreted as a position vector, can be understood as the representation of that variable in the emotion space. We take these representations of emotion variables from all label decoders, project them onto the unit sphere, and subject them to principal component analysis (PCA) \cite{Jolliffe02}. The first two principal components are shown in \fig\ref{fig:variables}.

As can be easily seen, equivalent emotion variables from different label formats have highly similar positions. For example, the representations for Anger in different label formats closely agree with each other (differently shaped markers overlap). This suggests that the technique for soft parameter sharing proposed in \S\ref{sec:mapping-training} works as intended. Also note that other proximities of high face-validity (Contempt being close to Anger; Joy being close to Valence) have emerged naturally from the data without explicit incentive.

\begin{table*}[t]
    \centering
    \caption{Emotion-Based Retrieval: Top Hits for English Word Queries in Different Content Types}
    \label{tab:retrieval}
    \begin{tabular}{|l|p{2cm}p{3.5cm}p{6cm}p{3cm}|}
    \hline
    \textbf{Query}     &  \textbf{English Words} & \textbf{Spanish Words} & \textbf{English Headlines} & \textbf{Images} \\
    \hline\hline
     rollercoaster    & infatuation\newline fireworks \newline lust & ascenso (promotion) \newline ganar (to win) \newline vencer (to overcome) & Baghdad plan is a success, Iraq prime minister tells Bush   & \raisebox{-.85\totalheight}{\includegraphics[width=3cm]{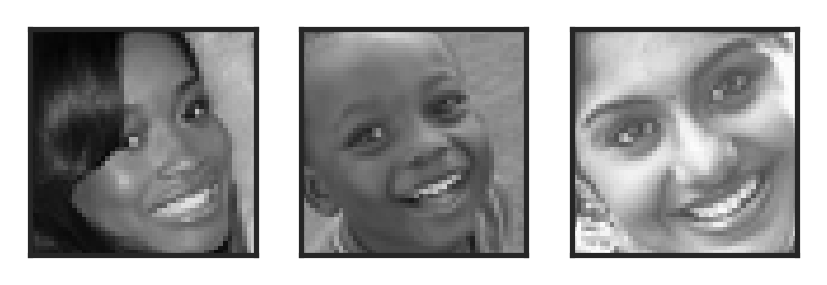}}\\
     \hline
     urine    & rancid\newline grime \newline fungus &  hediondo (foul) \newline tiña (ringworm) \newline vasallo (vassal) & Study: More kids exposed to online porn  & \raisebox{-.85\totalheight}{\includegraphics[width=3cm]{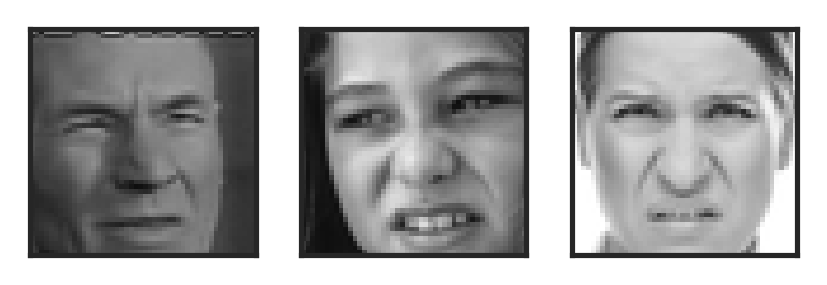}} \\
     \hline
     lonely    & depressed\newline burdened \newline sad & triste (sad) \newline posguerra (postwar) \newline tristeza (sadness) & Equipment on Plane in Brazil Collision May Have Been Faulty & \raisebox{-.85\totalheight}{\includegraphics[width=3cm]{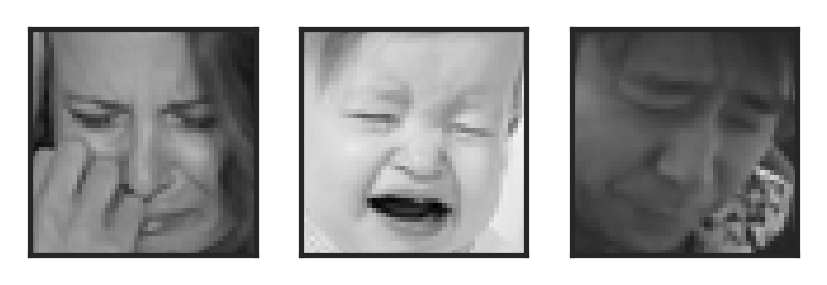}} \\
     \hline
     terrified    & scared\newline danger \newline fear & terrorismo (terrorism) \newline guerra (war) \newline asesinato (murder) & Iraqis fault delayed U.S. plan in attack  & \raisebox{-.85\totalheight}{\includegraphics[width=3cm]{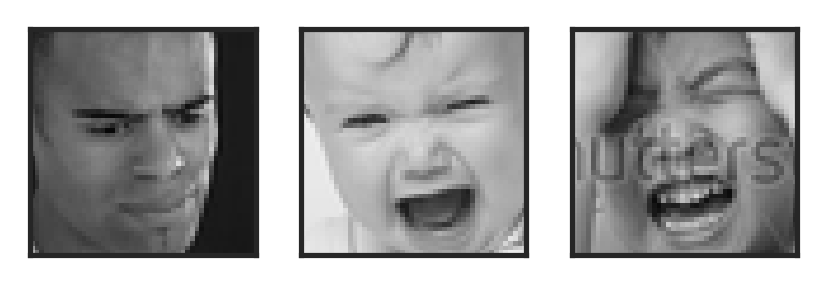}} \\
     \hline
     rage    & enraged\newline anger \newline angry & abuso (abuse) \newline chantaje (blackmail) \newline odio (hatred) & Cleric Said to Lose Reins Over Part of Iraqi Militia & \raisebox{-.85\totalheight}{\includegraphics[width=3cm]{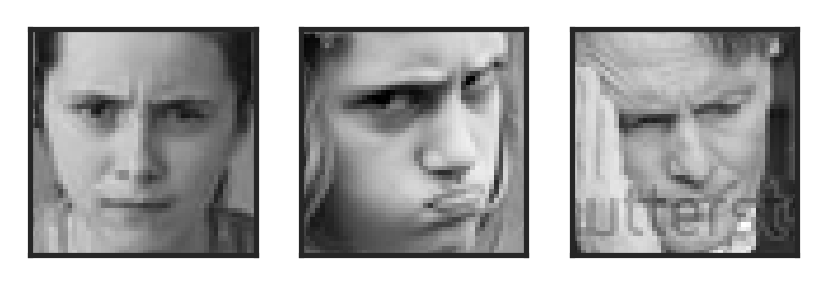}} \\
     \hline
    \end{tabular}
\end{table*}

As for the general structure of the space, we find that the first axis clearly differentiates between positive emotions (Valence, Joy, Surprise, and Dominance having having above-zero values) and negative ones (Fear, Sadness, Anger, Contempt, and Disgust having below-zero values), while the second axis seems to be linked to Arousal (Sadness having the highest value  whereas Fear, Arousal, and Surprise have the lowest ones).  Interestingly, this finding (which also was not directly incentivized by our method) agrees with the classic psychological result from behavioral experiments that Valence (positive vs. negative) and Arousal are the most important variables to explain variance in perceived emotion between stimuli \cite{Russell77}.

We point out that, once established, we can apply the same PCA transformation to any vector element of the emotion space, i.e., any embedding of any sample from any dataset employed above. Doing so for an illustrative selection of samples resulted in \fig\ref{fig:emotion-space} (see \S\ref{sec:intro}). As can be seen, the relative positioning of the samples and variables shows high face validity---samples associated with similar feelings appear close to each other as well as to their akin variable, independent of different languages, genres, modalities and media. 
Note, for instance, how the English words \textit{love} and \textit{sad} and their Spanish counterparts \textit{amor} and \textit{triste} are positioned very close to each other, respectively. Since our work did neither include any kind of machine translation component nor multilingual word embeddings, this result is \textit{solely} due to our shared embedding methodology, i.e., different content encoders assign similar emotion embeddings to these words because they evoke similar feelings in both languages.

\subsection{Cross-Modal Emotion-Based Retrieval}
\label{sec:emotion-retrieval}

Lastly, we want to illustrate the potential of our general emotion embeddings for retrieval-like applications. We computed and cached the emotion embeddings for each sample from each dataset using the respective content encoder trained in the supervised setting. Next, we computed a similarity matrix that stores the cosine similarity between each pair of samples independent of their domain, natural language or modality. This matrix can then be used to retrieve the \textit{emotionally closest neighbors in a given dataset} when queried accordingly. This feature serves as an illustration for a cross-modal retrieval application based on emotional similarity (see Table \ref{tab:retrieval}). We used words as exemplary queries because they are convenient and intuitive to enter for a potential user while also expressing fine emotional nuances, thus allowing to effectively search for content of a specific emotional tonality in large unstructured databases. A related use case of our emotion embeddings may also be "emotional indexing" where the feeling associated with a piece of content is summarized via automatically selected index terms. Such an application could also serve to describe the emotional load of an image to visually impaired users.

\section{Discussion \& Conclusions}\label{sec:discussion}

We presented an approach that learns stable and interoperable representations of emotion from heterogeneous datasets. Our solution consists of two steps: First, we train a \textit{Multi-Way Mapping Model} for the task of \enquote{translating} between different emotion label formats. While previous work for this task allowed to map between two sets of labels in \textit{one} direction, our model is the first to allow \textit{two-way} mappings between \textit{multiple} label formats. This feature is due to an encoder-decoder architecture with an intermediate layer shared between \textit{all} mapping directions that holds aforementioned emotion embeddings. 

In the second step, we make use of these emotion embeddings for different content types via \textit{Content Encoders}. With our newly proposed emotion label augmentation technique, we can use the Multi-Way Mapping Model as a teacher to train existing model architectures to predict emotion embeddings rather than labels for a piece of content, thus increasing the level of model abstraction at the representation layer. These embeddings can then be fed through the pre-processed decoder modules to yield predictions for a wide range of label formats. In our experiments, we have demonstrated for 16 datasets, including words and texts in different languages as well as images, that this approach widely increases the interoperability of existing emotion data without penalizing prediction quality. Further analyses provided evidence that the provisions we introduced worked as intended. In particular, our approach learns psychologically plausible and computationally valid representations of emotions in terms of emotion embeddings that are well suited to empower a range of novel applications, such emotion-based cross-modal and cross-lingual retrieval.

As far as limitations are concerned, it is important to point out that the two-step training process is rather complicated to apply. While it seems easy for other researchers to capitalize on the pre-trained Multi-Way Mapping Model as an off-the-shelves emotion embedding repository for their own experiments and applications, actually \textit{fitting} this model (say, for new emotion label formats or media formats such as audio or video streams) comes with a lot of effort of compiling the required mapping datasets. This is especially true because there is currently no way of \textit{adding} new label decoders to an existing mapping model. Thus, getting support for another label format requires training a new mapping model which, in turn, requires re-training of all content encoders.  More generally, we would like to point out that our proposed training process is \textit{one} way of fitting the collective encoder-decoder model, but it is unlikely to be the \textit{only} way how this can be done. Rather, we strongly encourage future work to devise new training methods for both the mapping model and the content encoders.

Another limitation of this work was the modest exploration of the hyperparameter space. The complexity of our experimental setup had us fall back on a rather simple strategy of hyperparameter selection---tuning the baseline model first on the respective dev set and then applying these  settings for our proposed model as well. The drawback of this approach is that it very likely leads to sub-optimal results for the latter model. It should thus be possible to outperform our already satisfactory performance with relative ease, simply by spending more time on any particular content decoder. We also found in our development experiments that the training parameters of the multi-way mapping model have a large impact on the down-stream performance. Hence, a detailed analysis of the impact of different hyperparameter settings would be appreciated. 

An important assumption of our work was that the quality of a content encoder would mainly depend on that of its respective base model, leading us to claim that the performance of our approach would "scale" with future developments in deep learning architectures. While our experimental results strongly suggest this to be true (in particular, by the small performance difference between baseline and proposed model in the supervised setting), it would really be  advisable to test this claim more directly. One way of doing so would be an experiment where the quality of the base model architecture is gradually degraded (e.g., successively removing blocks from a transformer model) and where the resulting performance is then correlated with that of the corresponding content encoder of the same architecture, ideally in a setting of automated hyperparameter selection.

Besides the domain and the label format, another source of heterogeneity of emotion data is its \enquote{viewpoint}, e.g., distinguishing between the emotion of the \textit{writer} of a piece of texts and its \textit{reader}, or whether a system should predict the emotion \textit{depicted} in a face image or which emotion viewing this image is likely to \textit{evoke}. These viewpoints lead to multiple---possibly diverging---labels for the same piece of content \cite{Buechel17eacl}. Even though not studied empirically,  we believe that our framework is already well-equipped to handle these sorts of perspectival effects (e.g., also for movies \cite{WangShangfei15,Muszynski21}). The solution we envision is to train distinct content encoders for each viewpoint, thus a piece of content can receive different positions within the emotion space and, by extension, different predicted labels, relative to the chosen encoder and viewpoint. Validating these considerations experimentally might also constitute a  focus of future work.

Finally, it might be rewarding to further extend our framework to additional modalities, such as audio and video data, or bio-signals like heart rate, skin conductivity, respiration or even brain imaging as physiological correlates of emotions. While we see no technical reason why our techniques should fail to generalize to such data as well, demonstrating that they actually do would indeed constitute a major step forward in the endeavor to unify research activities in affective computing.

\textbf{In conclusion}, our method for learning abstract emotion representations generalizes over highly heterogeneous data sources and offers techniques for leveraging these representations in downstream prediction problems. Together, these contributions significantly increase the interoperability, re-usability, and interpretability of emotion data, as our experiments on a wide range of datasets and our application example for emotion-based retrieval have shown. While it will be crucial to examine our framework both in more depth and breadth in future work, we believe this study offers a  reasonable starting point towards the methodological unification of affective computing methodologies and technologies.


\bibliographystyle{IEEEtran}
\bibliography{IEEEabrv,literature}

\end{document}